\documentclass[letterpaper]{article} % DO NOT CHANGE THIS
\usepackage{aaai2026}  % DO NOT CHANGE THIS
\usepackage{times}  % DO NOT CHANGE THIS
\usepackage{helvet}  % DO NOT CHANGE THIS
\usepackage{courier}  % DO NOT CHANGE THIS
\usepackage[hyphens]{url}  % DO NOT CHANGE THIS
\usepackage{graphicx} % DO NOT CHANGE THIS
\usepackage{natbib}  % DO NOT CHANGE THIS AND DO NOT ADD ANY OPTIONS TO IT
\usepackage{caption} % DO NOT CHANGE THIS AND DO NOT ADD ANY OPTIONS TO IT
\usepackage{algorithm}
\usepackage{algorithmic}

\usepackage{multirow}
\usepackage{colortbl}
\usepackage{tabularx, booktabs}
\usepackage{arydshln}
\usepackage{amsmath}
\usepackage{amssymb}
\usepackage{booktabs}
\usepackage{url}
\usepackage{amssymb}
\usepackage{bbding}
\usepackage{pifont}
\usepackage{wasysym}
\usepackage{utfsym}
\usepackage{fontawesome}
\usepackage{graphicx} 
\usepackage{caption} 
\usepackage{subcaption}

\definecolor{first}{RGB}{191,  232, 193} % 定义自己的颜色
\definecolor{second}{RGB}{225,  241,  170} % 定义自己的颜色
\definecolor{third}{RGB}{255, 251, 179} 

\usepackage{newfloat}
\usepackage{listings}
\DeclareCaptionStyle{ruled}{labelfont=normalfont,labelsep=colon,strut=off} % DO NOT CHANGE THIS
\floatstyle{ruled}
\newfloat{listing}{tb}{lst}{}
\floatname{listing}{Listing}
\title{PointSLAM++: Robust Dense Neural Gaussian Point Cloud-based
 SLAM}
\author{
    Xu Wang\textsuperscript{\rm 1}\equalcontrib, Boyao Han\textsuperscript{\rm 1}\equalcontrib, Xiaojun Chen\textsuperscript{\rm 1}, Ying Liu\textsuperscript{\rm 2}$^{\dagger}$,
Ruihui Li\textsuperscript{\rm 1}\thanks{Corresponding author.}
}
\affiliations{
    \textsuperscript{\rm 1}College of Computer Science and Electronic Engineering, Hunan University, China\\
    \textsuperscript{\rm 2}College of Information Science and Engineering, Hunan Normal University, China\\

\{xuwang0303, hby\_22, chenxiaojun\}@hnu.edu.cn, liu\_ying@hunnu.edu.cn, liruihui@hnu.edu.cn
}

\usepackage{bibentry}
\begin{document}

\maketitle

\vspace{-2.5em}

\begin{abstract}
% The ABSTRACT is to be in fully justified italicized text, at the top of the left-hand column, below the author and affiliation information.
% Use the word ``Abstract'' as the title, in 12-point Times, boldface type, centered relative to the column, initially capitalized.
% The abstract is to be in 10-point, single-spaced type.
% Leave two blank lines after the Abstract, then begin the main text.
% Look at previous \confName abstracts to get a feel for style and length.
Real-time 3D reconstruction is crucial for robotics and augmented reality, yet current simultaneous localization and mapping(SLAM) approaches often struggle to maintain structural consistency and robust pose estimation in the presence of depth noise. This work introduces PointSLAM++, a novel RGB-D SLAM system that leverages a hierarchically constrained neural Gaussian representation to preserve structural relationships while generating Gaussian primitives for scene mapping. It also employs progressive pose optimization to mitigate depth sensor noise, significantly enhancing localization accuracy. Furthermore, it utilizes a dynamic neural representation graph that adjusts the distribution of Gaussian nodes based on local geometric complexity, enabling the map to adapt to intricate scene details in real time. This combination yields high-precision 3D mapping and photorealistic scene rendering. Experimental results show PointSLAM++ outperforms existing 3DGS-based SLAM methods in reconstruction accuracy and rendering quality, demonstrating its advantages for large-scale AR and robotics.
\end{abstract}


\section{Introduction}
\label{sec:intro}

\begin{figure}[t]
  \centering
  % \maketitle
  \includegraphics[width=\linewidth]{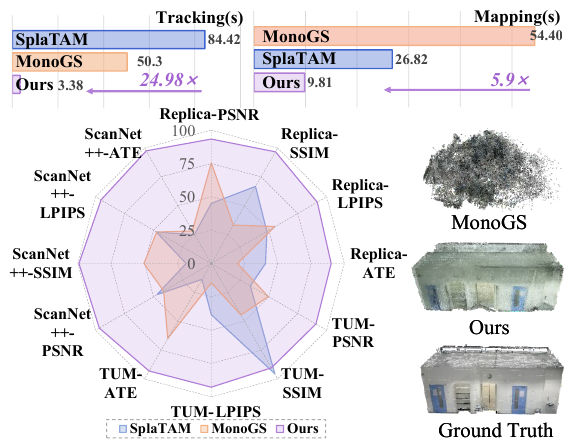}
  \caption{PointSLAM++ outperforms state-of-the-art methods from CVPR 2024 in photorealistic 3D reconstruction and camera tracking, excelling in key metrics. In complex environments where MonoGS fails, it maintains accurate localization and high-quality 3D mapping.}
  \label{fig:teaser}
\end{figure}

% Visual Simultaneous Localization and Mapping (SLAM) exploits monocular, stereo or RGB-D imagery to generate accurate 3D reconstructions and estimate camera trajectories. Its rapid uptake in robotics, virtual reality (VR) and augmented reality (AR) has fueled ever-greater demands for higher-fidelity rendering and more accurate, robust motion estimation~\cite{zubizarreta2020direct}. Traditional RGB-D SLAM (based on features or voxel) offers robust motion tracking but produces coarse details~\cite{du2011interactive,keller2013real,newcombe2011kinectfusion}, while implicit neural methods (e.g., iMAP~\cite{sucar2021imap}, NICE-SLAM~\cite{zhu2022nice}) deliver dense reconstructions at high computational cost and limited scale. This trade-off between efficiency and quality has driven the development of new SLAM representations~\cite{bylow2013real,canelhas2013sdf,dai2017bundlefusion,prisacariu1708infinitam,whelan2013robust}. Recently, 3D Gaussian Splatting (3DGS) has emerged as a hybrid model~\cite{kerbl20233d}, employing anisotropic Gaussian primitives and differentiable splatting to merge the efficiency of point clouds with the fidelity of neural rendering~\cite{zhu2024sni}. Systems like MonoGS~\cite{matsuki2024gaussian} (CVPR 2024) demonstrate real-time, photorealistic scene synthesis and joint pose–geometry optimization, yet challenges persist under noisy sensors, dynamic content, and very large-scale environments~\cite{hhuang2024photoslam,keetha2024splatam,matsuki2024gaussian,yan2024gs,hu2024cg}.
Visual SLAM uses monocular, stereo, or RGB-D images to reconstruct 3D scenes and estimate camera trajectories. Its widespread adoption in robotics, VR, and AR has intensified demands for higher-fidelity rendering and robust motion estimation~\cite{zubizarreta2020direct}. Traditional RGB-D SLAM (feature- or voxel-based) ensures stable tracking but yields coarse details~\cite{du2011interactive,keller2013real,newcombe2011kinectfusion}, while neural implicit methods (e.g., iMAP~\cite{sucar2021imap}, NICE-SLAM~\cite{zhu2022nice}) achieve dense reconstructions at high computational cost and limited scale. This efficiency–quality trade-off has spurred new SLAM representations~\cite{bylow2013real,canelhas2013sdf,dai2017bundlefusion,prisacariu1708infinitam,whelan2013robust}. Recently, 3D Gaussian Splatting (3DGS)~\cite{kerbl20233d} emerged as a hybrid paradigm, using anisotropic Gaussians and differentiable splatting to combine point-cloud efficiency with neural-rendering fidelity~\cite{zhu2024sni}. Systems like MonoGS~\cite{matsuki2024gaussian} demonstrate real-time photorealistic reconstruction and joint pose–geometry optimization, though challenges remain with sensor noise, dynamic scenes, and large-scale environments~\cite{hhuang2024photoslam,keetha2024splatam,matsuki2024gaussian,yan2024gs,hu2024cg}.

\begin{figure*}[t]
  \centering
  \includegraphics[width=\textwidth]{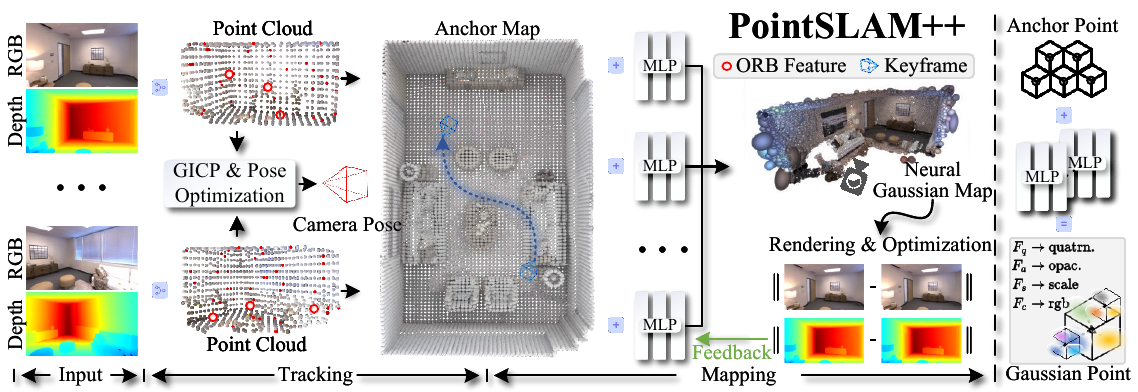}
  \caption{SLAM System Overview. The system’s input is a sequence of RGB-D frames. We generate a point cloud by downsampling and reprojecting the current depth image, and we estimate the current pose using GICP and ORB features. During tracking, we create anchor points from the point cloud and utilize a neural network to predict the Gaussian distribution in the scene, rendering the scene using a specialized Gaussian rasterizer. We continuously optimize the multi-layer perceptron (MLP) during the mapping process using the RGB-D information. Right: We use anchor points and features as input to the MLP to predict the various attributes of the Gaussians. }
  \label{fig:pipeline}
\end{figure*}

Current 3DGS-SLAM methods face three major challenges. First, reliance on precise depth information: an overdependence on high-quality depth maps causes the accuracy of tracking and reconstruction to fall in the presence of depth noise, missing data, or occlusions~\cite{chung2023orbeez,deng2024plgslam}. Second, Gaussian sphere overfitting: existing techniques fit a Gaussian sphere to every training view, ignoring local scene structure, which introduces significant redundancy and limits scalability in complex, large-scale environments~\cite{zhang2023hi}. Third, weak view dependence: view-dependent effects are baked into a single Gaussian parameter, yielding poor interpolation capabilities and low robustness to extensive viewpoint and illumination changes. Addressing these issues is essential for achieving scalable, high-fidelity 3DGS-SLAM.

This paper introduces PointSLAM++, a real-time 3D reconstruction system using RGB-D cameras. Using a learnable neural Gaussian representation with hierarchical constraints and progressive pose optimization, it enables precise mapping and realistic rendering of scenes.

To eliminate reliance on precise depth information, we propose a progressive optimization scheme to achieve accurate pose estimation: we apply the iterative closest point (ICP) algorithm for rigid registration of depth data and integrate ORB-feature-based visual constraints for joint refinement. Within a multi-scale pyramid, we simultaneously minimize the depth point-cloud residuals and the ORB-based visual reprojection errors~\cite{campos2021orb}. However, under challenging conditions—such as high-speed motion, low-texture regions, or dynamic interference—tracking can still be lost. To recover the camera pose quickly when conventional tracking fails, we introduce a relocalization mechanism based on global feature matching.

To meet the demands of real-time incremental mapping and high-precision reconstruction, stable ORB feature point clouds from tracking are fused with depth sensor data for map construction. The point cloud is then converted into neural Gaussian “anchors” endowed with position, scale, and related parameters, and their distributions are constrained by the scene’s geometric structure. A two-tier anchor mechanism includes primary anchors for global structure and secondary anchors for fine detail, enabling incremental updates and detail enrichment while preserving robust pose estimation and geometric accuracy. Finally, camera-view vectors are embedded into each anchor’s appearance features to compensate for viewpoint-dependent lighting and reflectance effects, thereby enhancing reconstruction quality under complex illumination and varying viewpoints.

In summary, our contributions are as follows:
\begin{itemize}
    \item We present PointSLAM++, a real-time RGB-D SLAM system that unifies geometric and visual pose refinement with a hierarchical Neural-Gaussian representation for precise mapping and photorealistic rendering.

    \item We develop a progressive multi-scale pose optimization that jointly minimizes ICP depth residuals and ORB reprojection errors, and we add a global-feature relocalization module to recover from fast motion or low-texture tracking failures.
    
    \item We propose a two-tier neural-Gaussian anchor framework that fuses ORB feature clouds with depth data and embeds camera-view vectors to adaptively enhance detail and compensate for view-dependent lighting.

\end{itemize}
\section{Related Work}
\label{sec:relatedWork}

%-------------------------------------------------------------------------
\subsection{Classical RGBD SLAM}

% Traditional RGB-D dense SLAM frameworks optimize 3D reconstruction and localization using geometric primitives, evolving in three stages: KinectFusion~\cite{newcombe2011kinectfusion} established real-time reconstruction with TSDF and ICP~\cite{gelfand2003geometrically}; ElasticFusion~\cite{whelan2016elasticfusion} improved topological consistency in large-scale scenes with graph-optimized non-rigid deformation; and BundleFusion~\cite{dai2017bundlefusion} addressed robustness in large environments through hierarchical optimization and global loop closure. Voxel and point-based semantic SLAM systems are computationally intensive and struggle with consistency. Meanwhile, SDF-based methods, though geometrically precise, face severe memory and scalability issues due to uniform grids~\cite{wang2022hf}. Additionally, post-processing methods, such as conditional random field segmentation, lead to label misalignment and boundary ambiguity due to decoupled optimization.

Traditional RGB-D dense SLAM evolves through three stages: KinectFusion~\cite{newcombe2011kinectfusion} achieved real-time TSDF reconstruction via ICP~\cite{gelfand2003geometrically}; ElasticFusion~\cite{whelan2016elasticfusion} enhanced large-scale topological consistency through graph-based non-rigid deformation; and BundleFusion~\cite{dai2017bundlefusion} improved robustness via hierarchical optimization and global loop closure. However, voxel and point-based semantic SLAM remain computationally heavy and inconsistent. SDF-based methods, while accurate, suffer from memory and scalability limits due to uniform grids~\cite{wang2022hf}. Moreover, post-processing like CRF segmentation causes label misalignment and blurred boundaries from decoupled optimization.

\begin{figure}[t]
    \centering
    \includegraphics[width=1\linewidth]{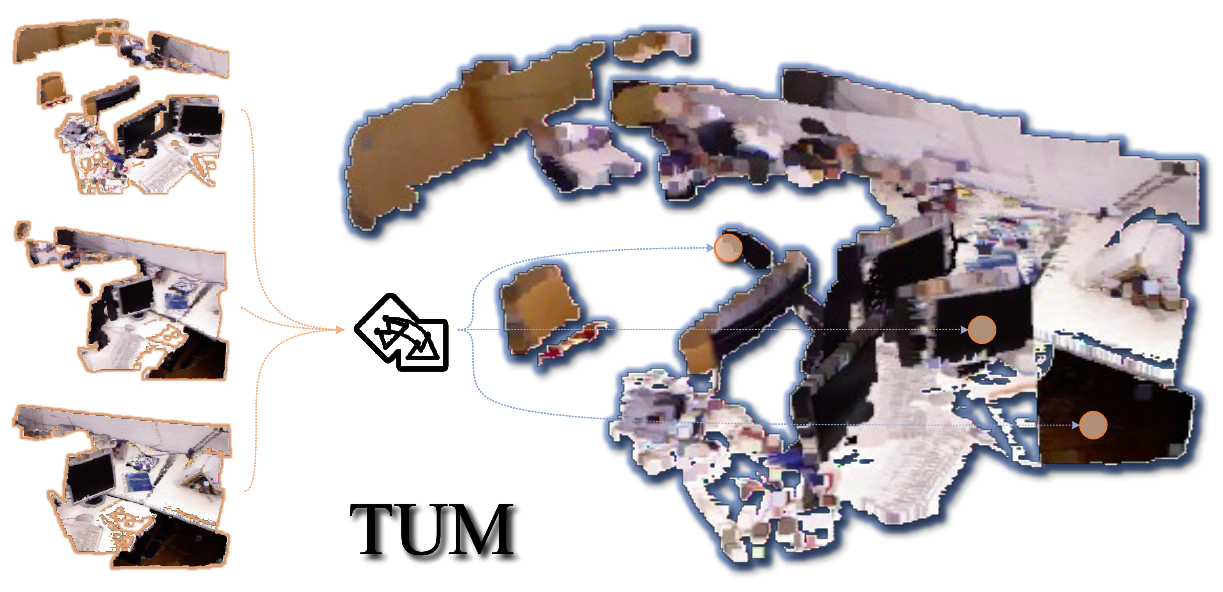}
    \caption{GICP Mismatches in TUM Scenes. Even with high-precision depth data, real-world noise can lead to misalignments in GICP-based pose estimation, as illustrated here in the TUM dataset.} 
    %The enlarged partial view in the figure is convenient for comparison. }
    \label{fig:icp}
\end{figure}

%-------------------------------------------------------------------------
\subsection{NeRF-based RGBD SLAM}

% Recently, the exceptional performance of Neural Radiance Fields~\cite{mildenhall2020nerf} (NeRF) in scene representation has sparked significant interest in utilizing NeRF for 3D reconstruction and localization. The core of NeRF-based SLAM lies in constructing continuous implicit functions through MLPs, which map 3D spatial coordinates to radiance brightness and volume density distributions~\cite{wang2023co,johari2023eslam}. This approach transcends the resolution limitations of traditional voxel or point cloud representations, demonstrating sub-voxel reconstruction precision in static scenes (as seen in iMAP~\cite{sucar2021imap} and NICE-SLAM~\cite{zhu2022nice}). However, these methods still face bottlenecks in optimization efficiency of implicit representations, dynamic scene disentanglement capabilities, and scalability to large environments, which limit their robustness in real-world open environments~\cite{jiang2023alignerf}.

The strong scene representation ability of Neural Radiance Fields~\cite{mildenhall2020nerf} (NeRF) has driven their use in 3D reconstruction and localization. NeRF-based SLAM builds continuous implicit functions via MLPs that map 3D coordinates to radiance and density~\cite{wang2023co,johari2023eslam}. This overcomes the resolution limits of voxel and point-based models, achieving sub-voxel accuracy in static scenes (e.g., iMAP~\cite{sucar2021imap}, NICE-SLAM~\cite{zhu2022nice}). Yet, challenges remain in optimization efficiency, dynamic scene disentanglement, and scalability, hindering robustness in large, real-world environments~\cite{jiang2023alignerf}.

%-------------------------------------------------------------------------
\subsection{Gaussian-based RGBD SLAM}
% RGB-D SLAM based on 3D Gaussian Splatting~\cite{kerbl20233d} has recently revolutionized the field. Methods like Gaussian-Splatting SLAM ~\cite{matsuki2024gaussian,liso2024loopy} leverage analytical Jacobian matrices and geometric regularization to achieve photorealistic reconstructions, outperforming neural radiance fields (NeRF) in accuracy. Another key approach, GS-ICP SLAM~\cite{ha2024rgbd}, incorporates Gaussian covariance into a generalized iterative closest point (G-ICP) framework, enabling sub-millimeter alignment at twice the speed of traditional ICP. However, challenges remain, such as depth sensor noise biasing covariance initialization and requiring intensive regularization, as well as dynamic scene modeling constrained by offline motion segmentation or overly simplistic assumptions for low-dynamic environments~\cite{zhu2024loopsplat,mao2024ngel}. Additionally, incremental optimization often causes misalignments in large-scale reconstructions. Overcoming these challenges will demand improvements in noise-resilient regularization and scalable dynamic scene modeling for practical SLAM applications.

3D Gaussian Splatting~\cite{kerbl20233d} has recently reshaped RGB-D SLAM. Approaches such as Gaussian-Splatting SLAM~\cite{matsuki2024gaussian,liso2024loopy} exploit analytical Jacobians and geometric regularization for photorealistic reconstruction, surpassing NeRF in accuracy. GS-ICP SLAM~\cite{ha2024rgbd} further integrates Gaussian covariance into a G-ICP framework, achieving sub-millimeter alignment at twice the speed of conventional ICP. Remaining challenges include depth noise affecting covariance initialization, the need for heavy regularization, and limited dynamic scene modeling due to offline segmentation or simplified motion assumptions~\cite{zhu2024loopsplat,mao2024ngel}. Incremental optimization also causes misalignment in large-scale mapping. Addressing these issues requires noise-robust regularization and scalable dynamic modeling for practical SLAM deployment.
\section{Method}
\label{sec:method}

An overview of our reconstruction pipeline is shown in Fig~\ref{fig:pipeline}. In Section 3.1, we first introduce the progressive pose optimization(PPO) framework . Then, in Section 3.2, we describe the neural Gaussian representation and its density-adaptive optimization strategy. Finally, in Section 3.3, we provide a detailed description of the entire online reconstruction process based on neural Gaussian representation and progressive pose optimization.

\subsection{Progressive Pose Optimization}

In the camera pose localization problem, the primary objective is to accurately estimate camera trajectories within a global coordinate system. To achieve highly robust pose localization, we propose a progressive optimization framework that systematically refines pose estimates through cascaded stages, integrating point-cloud registration with feature-based refinement to mitigate common sources of error in real-world scenarios.

We begin with Generalized Iterative Closest Point (GICP) as the front-end odometry for coarse pose estimation. Given the local point cloud $\mathcal{P} = \{ p_i \mid i = 1, 2, \dots, N \}$ from the previous frame and $\mathcal{Q} = \{ q_i \mid i = 1, 2, \dots, N \}$ from the current frame, the optimal transformation $\mathbf{T} \in SE(3)$ minimizes the registration error:

\begin{equation}
    \mathbf{T}^* = \arg\min_{\mathbf{T}} \sum_{j=1}^{N} \sum_{i \in C(j)} d(\mathbf{T} \cdot q_j, p_i)^2
\end{equation}

\noindent where $C(j)$ denotes the correspondence set for point $q_j$, and $d(\cdot)$ is the distance metric. GICP iteratively optimizes correspondences and the transformation to yield an initial pose estimate.

Although accurate depth improves GICP, real-world noise often causes mismatches (Fig.~\ref{fig:icp}). To mitigate this, we exploit the robustness of ORB features with more reliable depths. Extracted via multi-scale pyramids and fused with depth data, these features form local point clouds registered to the global map through point-to-plane constraints, initialized by the coarse GICP pose $\mathbf{T}$ for faster convergence and higher accuracy.

% While high-precision depth data can enhance GICP's accuracy, noisy real-world measurements often induce mismatches (Fig.~\ref{fig:icp}). To counteract this, we leverage the robustness of ORB features, whose associated depths are typically more reliable. ORB features are extracted via multi-scale pyramids and fused with depth information to create local feature point clouds. These are then registered against the global map using point-to-plane distance constraints, seeded with the coarse GICP pose $\mathbf{T}$, thereby accelerating convergence and boosting precision.

Building on this foundation, our framework further refines the pose by minimizing reprojection errors through bundle adjustment, which jointly optimizes camera poses and 3D landmarks. This progressive strategy—unique in its seamless fusion of odometry, feature-driven registration, and optimization—employs GICP+ORB for initialization, anchors the first frame to resolve scale ambiguity, parameterizes poses on the $SE(3)$ manifold, and utilizes a robust Levenberg-Marquardt solver with Huber loss and depth priors for stable convergence. The core objective unifies these elements:

\begin{equation}
    \min_{\xi} \sum_{k=1}^{K} \rho\left( e_k^T \Sigma_k^{-1} e_k \right)
\end{equation}

\noindent where:
- $\xi \in \mathfrak{se}(3)$ parameterizes the pose $T = \exp(\xi^\wedge) \in SE(3)$,
- $K$ is the number of observed features,
- $e_k = x_k - \pi(K T X_k) \in \mathbb{R}^2$ denotes the reprojection error, with $x_k$ as observed pixels, $\pi$ as the projection function, $K$ as intrinsics, and $X_k$ as 3D points,
- $\Sigma_k^{-1} \in \mathbb{R}^{2 \times 2}$ is the information matrix,
- $\rho$ is a robust kernel for outlier rejection.

\subsubsection{Pose Recovery and Optimization.} In challenging conditions such as high-speed motion, blur, sparse textures, or dynamic scenes, tracking may falter. Our recovery mechanism innovatively combines feature matching with established techniques to restore poses reliably. An initial correspondence set $M$ is formed via descriptor matching between current features and map points. The Perspective-n-Point (PnP) solver, embedded in a RANSAC loop, then computes a robust initial pose by minimizing reprojection errors as per Equation (2), incorporating the robust kernel $\rho$ to handle mismatches.

Subsequent nonlinear optimization refines this estimate using the same Lie algebra formulation and objective from Equation (2). This layered architecture—coarse GICP, ORB-augmented registration, bundle adjustment, and PnP+RANSAC recovery—sets our method apart by offering adaptive robustness that surpasses the isolated application of these components in existing pose tracking systems. Relocalization succeeds with high confidence when inlier rates exceed a predefined threshold, bolstering overall system resilience in complex environments.

\begin{figure}[t]
    \centering
    \includegraphics[width=1\linewidth]{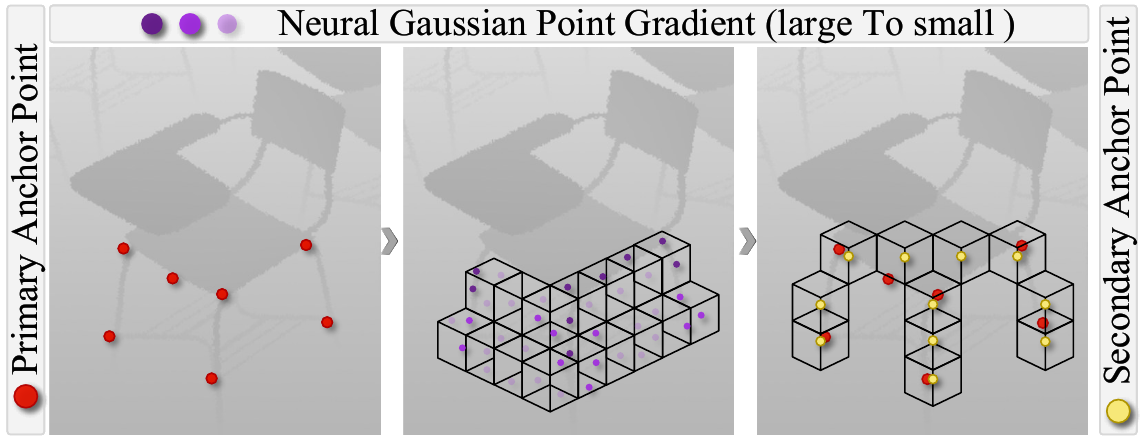}
    \caption{\textbf{Hierarchical Anchor Point Optimization.} Primary anchor points (red) are derived from ORB features and remain stable throughout SLAM, while secondary anchor points (yellow) are introduced or culled based on voxel-wise gradient checks (purple) of neural Gaussian points.} 
    %The enlarged partial view in the figure is convenient for comparison. }
    \label{fig:anchor}
\end{figure}

\subsection{Neural-Gaussian Representation}

High-quality 3D reconstruction relies on accurate map initialization. While COLMAP-generated sparse point clouds serve as effective priors, they are not suitable for SLAM applications. To address this, we propose leveraging stable ORB feature point clouds extracted during SLAM tracking, supplemented by depth sensor data, for efficient and lightweight initialization.

At startup, the mapping thread seeds the global map with the first frame’s ORB feature point cloud. It then incrementally incorporates feature points from each subsequent keyframe, continuously updating and expanding the map. To achieve high-fidelity scene reconstruction, we convert traditional ORB point clouds into neural point clouds with enhanced representational power:
\begin{equation}
N_v = \{ (p_v, f_v, l_v, O_v) \}
\end{equation}
Each neural Gaussian point consists of the center position $p_v \in \mathbb{R}^3$ of a voxel $v$, local context features $f_v \in \mathbb{R}^{32}$, scaling factors $l_v \in \mathbb{R}^3$, and $k$ learnable offsets $O_v \in \mathbb{R}^{k \times 3}$. Similar to the Scaffold-GS method~\cite{scaffoldgs}, we refer to these points as "anchor points".

Each anchor point generates $K$ neural Gaussian models through a Multi-Layer Perceptron (MLP). Each neural Gaussian model is defined as:
\begin{equation}
G = \{ (\mu, \alpha, q, s, c) \}
\end{equation}
\noindent where $\mu \in \mathbb{R}^3$ is the 3D position, $\alpha \in \mathbb{R}$ is the opacity, $q \in \mathbb{R}^4$ is a quaternion controlling the covariance, $s \in \mathbb{R}^3$ is a scaling vector, and $c \in \mathbb{R}^3$ is the color.

Specifically, the relationship between the 3D position of a neural Gaussian model and its learnable offset is given by:
\begin{equation}
\{\mu_0, \ldots, \mu_{k-1}\} = x_v + \{O_0, \ldots, O_{k-1}\} \cdot l_v 
\end{equation}
\noindent where $\{O_0, O_1, \ldots, O_{k-1}\} \in \mathbb{R}^{k \times 3}$ represents a set of learnable offsets and $l_v$ is the scaling factor associated with the anchor point.

The set of attributes, $A$, for the $k$ neural Gaussians generated by each anchor point is computed by an MLP network $F_a$, as follows:
\begin{equation}
\{A_0, \ldots, A_{k-1}\} = F_a(\hat{f}_v, \delta_v c, \vec{d}_v c) 
\end{equation}
\noindent where $A$ represents any attribute of the neural Gaussian (opacity $\alpha$, color $c$, quaternion $q$, or scaling $s$). $\{A_0, \ldots, A_{k-1}\}$ denotes the set of that specific attribute for the $k$ neural Gaussians generated by a single anchor point. Subsequently, rendering is performed using a Gaussian renderer. $F_a$ represents the MLP network used to compute the specific attribute $A$. $\delta_v c$ and $\vec{d}_v c$ represent the relative viewing distance and direction between the camera and the anchor point. The specific calculation details will be provided in the supplementary materials.

\subsubsection{Hierarchical Anchor Point Optimization.} Unlike point clouds initialized using SfM reconstruction, the map maintained by SLAM algorithms is a dynamic, incremental representation. The Mapping and Tracking threads share the same map before the Tracking thread terminates. Therefore, extensive modifications to anchor points can directly impact camera pose estimation.

To address this, we introduce hierarchical anchor points. Anchor points generated from ORB features are designated as primary anchor points, which cannot undergo splitting or deletion and are optimized using a small learning rate. As ORB features have limited modeling capabilities in texture-scarce regions, we introduce secondary anchor points generated from depth sensor data and smaller voxels. Both primary and secondary anchors are collectively denoted as $N_v$.

To determine where secondary anchor points are needed, we construct voxels of size $\epsilon_g$ to spatially quantize neural Gaussian points. For each voxel, we calculate the average gradient $\nabla g$ over $N$ training iterations. If $\nabla g > \tau_g$ and no anchor point exists in the voxel, a new secondary anchor point is deployed using depth information at the voxel center. If a secondary anchor point already exists, its opacity is halved until it falls below the threshold and is culled.

\subsubsection{View-Dependent Environment Compensation.} During camera motion, scene appearance varies due to viewpoint changes. These variations result from ambient lighting changes affecting camera exposure and material properties like specular reflections causing view-dependent radiance. These factors require neural Gaussians at the same location to render different pixel values from different viewpoints. However, traditional Gaussian models with their isotropic assumption cannot adequately handle these view-dependent appearance changes, even with spherical harmonics.

To address this limitation, we encode the camera viewing direction as a unit vector $\mathbf{v}\in\mathbb{R}^3$ and embed it together with the anchor appearance feature $\hat{\mathbf{f}}_a$ into the contextual features:

\begin{equation}
\mathbf{f}_a' = \mathrm{Embed}(\hat{\mathbf{f}}_a,\;\mathbf{v})
\end{equation}
where $\hat{\mathbf{f}}_a$ is the anchor appearance feature (i.e.\ the RGB color), $\mathbf{v}$ is the normalized view‐direction vector, and $\mathrm{Embed}(\cdot)$ is implemented by an MLP. The output $\mathbf{f}_a'$ is the view‐direction‐enhanced contextual feature, which serves as the initial form of the contextual feature $\mathbf{f}_v$ and is then continuously updated during training. Hence, we denote this initial contextual feature by $\mathbf{f}_a$. This approach explicitly integrates view direction with appearance features and overcomes traditional methods’ dynamic‐viewpoint limitations.

\subsection{SLAM System}

We have developed a real-time SLAM system based on a neural Gaussian representation and a progressive pose optimization framework.  Assume we have a map represented by a set of 3D Gaussians, constructed from previous camera frames 1 to $t$.  For a new RGB-D frame $t+1$, our SLAM system performs the following steps:

\begin{itemize}
    \item \textbf{Camera Tracking:} We estimate the camera pose by minimizing the reprojection error and depth error between the current frame and the global map, utilizing the image and depth information from the current RGB-D frame $t+1$.  A coarse estimate is first obtained using GICP, followed by optimization using point cloud constraints based on ORB features and the reprojection error.

    \item \textbf{Gaussian Densification:} When the current frame $t+1$ is determined to be a keyframe, we add new anchor points to the map based on the ORB feature points and their corresponding depth information from that frame. This enhances the map's ability to represent fine details.

    \item \textbf{Map Update:} If the current frame $t+1$ is selected as a keyframe, we update the parameters of all neural Gaussians in the map by minimizing the RGB and depth errors on the current keyframe, as well as on historical keyframes that have overlap with the current keyframe. If the current frame is not a keyframe, a subset of historical keyframes is randomly selected for optimization to maintain continuous map refinement and consistency.
\end{itemize}

\subsubsection{Keyframe Selection.} We use a dynamic keyframe selection strategy, similar to GS-ICP SLAM~\cite{ha2024rgbd}, based on geometric consistency.  A new keyframe is selected if the average point cloud distance between the current frame and the map exceeds a threshold, or if too few points fall below that threshold.  To mitigate error accumulation, new Gaussian points are only added to non-overlapping regions of the new keyframe.

\subsubsection{Map Update.} During anchor point optimization, we employ multiple losses—color, SSIM (based on 3D Gaussian splatting), and geometric—to ensure color fidelity, structural consistency, and geometric coherence in rendered images. A scale regularization term further guides Gaussian ellipsoids toward isotropic spheres, reducing anisotropy and balancing tracking accuracy with rendering realism.
% When optimizing the anchor point parameters, we incorporate multiple loss functions, including the color loss and Structural Similarity (SSIM) loss based on the original 3D Gaussian splatting, as well as a geometric loss. This ensures the rendered images exhibit consistency in color, structural integrity, and spatial geometric features. Furthermore, we introduce a scale regularization term that encourages the Gaussian ellipsoids to converge towards isotropic Gaussian spheres. This strategy effectively reduces anisotropy, thereby achieving a balance between trajectory tracking accuracy and rendering realism.

\begin{equation}
\begin{aligned}
\mathcal{L}_{mapping} & = \lambda_{I_1} \mathcal{L}_1\left(I, I_{gt}\right) + \lambda_{I_2} \mathcal{L}_{D\text{-}SSIM}\left(I, I_{gt}\right) \\
& + \lambda_D \mathcal{L}_1\left(D, D_{gt}\right)
\end{aligned}
\end{equation}

\begin{figure*}[t!]
  \centering
  \includegraphics[width=\textwidth]{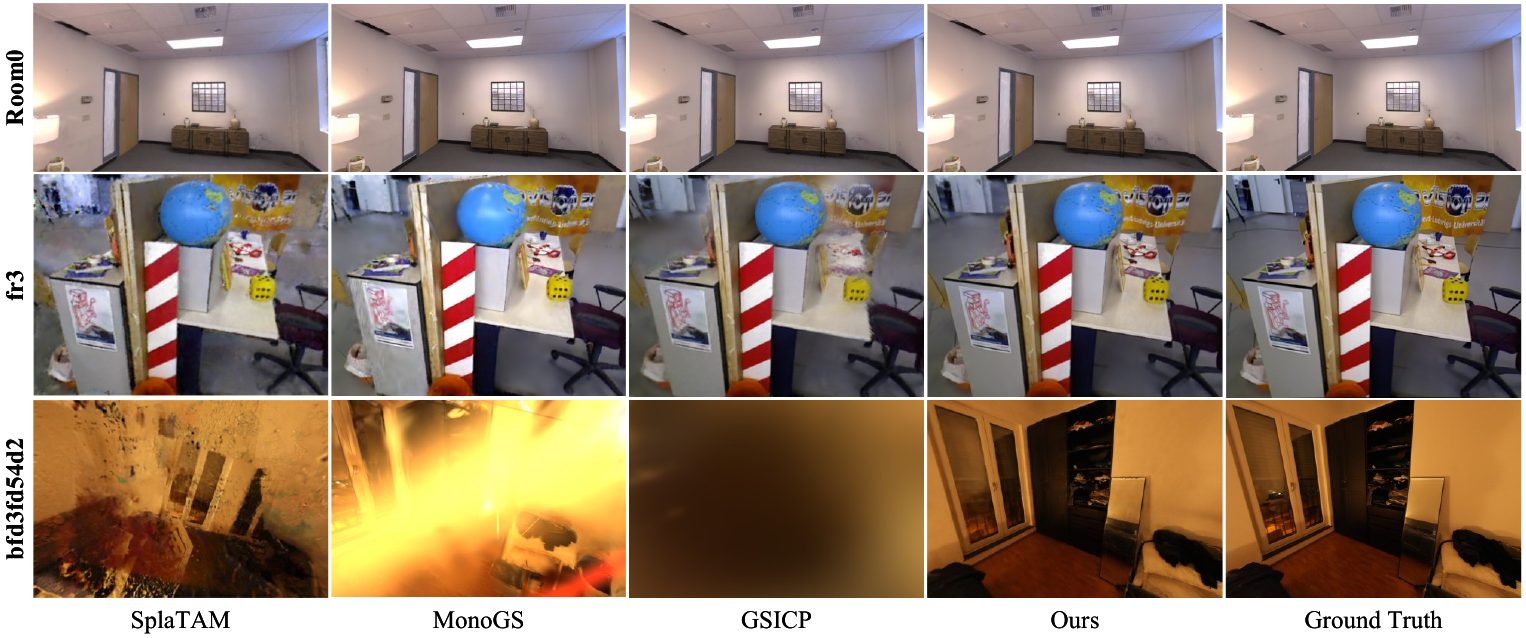}
  \caption{\textbf{Comparison of Rendering Results.} We present scenes from three datasets: Room0 (Replica), fr3 (TUM-RGBD), and bfd3fd54d2 (ScanNet++). Our method achieves rendering results that closely match the ground truth, demonstrating high accuracy. Additional results are available in the supplementary materials.}
  \label{fig:comp}
\end{figure*}

\section{Experiments}
\label{sec:experiments}

\subsection{Experimental Setup}

\subsubsection{Implementation Details.} Our SLAM system is implemented on a desktop computer equipped with an Intel Xeon Silver 4314 CPU and an Nvidia RTX 3090 GPU. The mapping and tracking components are implemented in Python and built upon the Pytorch framework.

\subsubsection{Datasets.} We evaluate our method on three datasets. Replica~\cite{replica19arxiv} provides high-precision synthetic RGB-D images suitable for basic validation. TUM-RGBD dataset~\cite{sturm12iros}, with precise camera poses from motion capture systems, is standard for SLAM tracking accuracy evaluation despite its low image quality and noisy depth data. ScanNet++~\cite{yeshwanth2023scannet++} is a large-scale dataset with both high and ordinary quality indoor scene data, featuring larger camera pose intervals and depth map errors, allowing us to test system robustness under fast camera movement or sparse texture conditions.

\subsubsection{Baselines.} We compare our method with state-of-the-art RGB-D SLAM approaches, including NICE-SLAM~\cite{Zhu2022CVPR}, Point-SLAM~\cite{Sandström2023ICCV}, and concurrent Gaussian SLAM methods such as SplaTAM~\cite{keetha2024splatam}, MonoGS~\cite{matsuki2024gaussian}, Photo-SLAM~\cite{hhuang2024photoslam}, and GS-ICP SLAM~\cite{ha2024rgbd}. To ensure the credibility of the results, we reproduce the experiments using the official code provided by each method. For the ScanNet++ dataset~\cite{yeshwanth2023scannet++}, due to its sparse viewpoint distribution, we double the scene (i.e., the map in SLAM) update frequency for all methods to ensure a fair comparison.

\begin{table*}[t]
\renewcommand\arraystretch{1}
  \centering
  \resizebox{\linewidth}{!}{
    \begin{tabular}{lcccccccccc}
    \toprule
    \multirow{2}[2]{*}{Method} & & \multicolumn{4}{c}{Replica} & & \multicolumn{4}{c}{ScanNet++} \\
    \cmidrule{3-6} \cmidrule{8-11}
    & & PSNR$\uparrow$ & SSIM$\uparrow$ & LPIPS$\downarrow$ & ATE RMSE$\downarrow$ & & PSNR$\uparrow$ & SSIM$\uparrow$ & LPIPS$\downarrow$ & ATE RMSE$\downarrow$ \\
    \midrule
    {NICE-SLAM~\cite{zhu2022nice}}  & $\vert$ & 24.42 & 0.809 & 0.233 & 1.95 & $\vert$ & - & - & - & - \\
    {Point-SLAM~\cite{Sandström2023ICCV}} & $\vert$ & 5.17 & \underline{0.975} & 0.124 & 0.52 & $\vert$ & - & - & - & -\\
    {SplaTAM~\cite{keetha2024splatam}} & $\vert$ & 34.11 & 0.970 & 0.100 & 0.36 & $\vert$ & 14.32 & 0.482 & 0.599 & 287.36\\
    {MonoGS~\cite{matsuki2024gaussian}} & $\vert$ & 37.5 & 0.960 & 0.070 & 0.58 & $\vert$ & 12.90 & 0.648 & 0.603 & \underline{101.19} \\
    {Photo-SLAM~\cite{hhuang2024photoslam}} & $\vert$ & 34.96 & 0.942 & 0.059 & 0.61 & $\vert$ & \ding{53} & \ding{53} & \ding{53} & \ding{53}\\
    {GS-ICP SLAM~\cite{ha2024rgbd}} & $\vert$ & \underline{38.83} & \underline{0.975} & \underline{0.041} & \textbf{0.16} & $\vert$ & \underline{14.94} & \underline{0.776} & \underline{0.446} & 111.37 \\
    \textcolor[rgb]{ .122,  .137,  .157}{Ours} & $\vert$ & \textbf{39.46} & \textbf{0.979} & \textbf{0.027} & \underline{0.19}  & $\vert$ & \textbf{26.51} & \textbf{0.905} & \textbf{0.148} & \textbf{6.73}\\
    \bottomrule
    \end{tabular}}%
     \caption{Quantitative evaluation of ours compared to existing methods for RGB-D cameras on Replica~\cite{replica19arxiv} and ScanNet++~\cite{yeshwanth2023scannet++} datasets, where '–' denotes systems failing to provide valid results and '×' indicates systems unable to complete tracking or reconstructio. More quantitative results can be found in the supply materials.}
  \label{tab:rep&sca}%
\end{table*}%

\begin{table}[t]
    \centering
    \small
    \resizebox{\linewidth}{!}{
    \begin{tabular}{lcccccc}
    \toprule[0.6pt]
        Method & Metric & fr1/desk & fr2/xyz & fr3/office & Avg. \\ \midrule[0.5pt]
\multirow{3}{*}{NICE-SLAM}
      & PSNR$\uparrow$ & 13.83 & 17.87 & 12.89 & 14.86 \\
      & SSIM$\uparrow$ & 0.569 & 0.718 & 0.554 & 0.614 \\
      & LPIPS$\downarrow$ & 0.482 & 0.344 & 0.498 & 0.441 \\ \midrule[0.5pt]
\multirow{3}{*}{Point-SLAM}
      & PSNR$\uparrow$ & 13.87 & 17.56 & 18.43 & 16.62 \\
      & SSIM$\uparrow$ & 0.627 & 0.708 & 0.754 & 0.696 \\
      & LPIPS$\downarrow$ & 0.546 & 0.585 & 0.448 & 0.526 \\ \midrule[0.5pt]
\multirow{3}{*}{SplaTAM}
      & PSNR$\uparrow$ & 22.00 & 24.50 & 21.90 & 22.80 \\
      & SSIM$\uparrow$ & \underline{0.860} & \textbf{0.950} & \underline{0.880} & \textbf{0.897} \\
      & LPIPS$\downarrow$ & \underline{0.230} & \underline{0.100} & 0.200 & 0.177 \\ \midrule[0.5pt]
\multirow{3}{*}{MonoGS}
      & PSNR$\uparrow$ & \underline{23.69} & \underline{24.68} & \underline{22.83} & \underline{23.7329} \\
      & SSIM$\uparrow$ & 0.786 & 0.793 & 0.770 & 0.783 \\
      & LPIPS$\downarrow$ & 0.245 & 0.225 & 0.277 & 0.249 \\ \midrule[0.5pt]
\multirow{3}{*}{Photo-SLAM}
      & PSNR$\uparrow$ & 20.87 & 22.10 & 22.74 & 21.90 \\
      & SSIM$\uparrow$ & 0.743 & 0.765 & 0.780 & 0.763 \\
      & LPIPS$\downarrow$ & 0.239 & 0.169 & \underline{0.145} & 0.184 \\ \midrule[0.5pt]
\multirow{3}{*}{GS-ICP SLAM}
      & PSNR$\uparrow$ & 17.95 & 23.13 & 20.50 & 20.53 \\
      & SSIM$\uparrow$ & 0.710 & 0.829 & 0.758 & 0.766 \\
      & LPIPS$\downarrow$ & 0.296 & 0.141 & 0.232 & 0.223 \\ \midrule[0.5pt]
\multirow{3}{*}{Ours}
      & PSNR$\uparrow$ & \textbf{25.05} & \textbf{27.11} & \textbf{26.33} & \textbf{26.16} \\
      & SSIM$\uparrow$ & \textbf{0.876} & \underline{0.899} & \textbf{0.881} & \underline{0.885} \\
      & LPIPS$\downarrow$ & \textbf{0.123} & \textbf{0.078} & \textbf{0.118} & \textbf{0.106} \\ \bottomrule[1pt]
    \end{tabular}}
    \caption{Rendering Performance on TUM-RGBD~\cite{sturm12iros}. } 
\label{tab:tum-r}
\end{table}

\begin{table}[t]
    \centering
    \small
    \resizebox{\linewidth}{!}{
    \begin{tabular}{lccccc}
    \toprule[0.6pt]
        Method & fr1/desk & fr2/xyz & fr3/office & Avg. \\ \midrule[0.5pt]
NICE-SLAM & 4.26 & 31.73 & 3.87 & 13.29 \\
Point-SLAM & 4.34 & 1.31 & 3.48 & 3.04 \\
SplaTAM & 3.35 & 1.24 & 5.16 & 3.25 \\
MonoGS & \textbf{1.50} & 1.44 & 1.49 & 1.48 \\
Photo-SLAM & 2.60 & \underline{0.35} & \textbf{1.00} & \underline{1.32} \\
GS-ICP SLAM & 3.50 & 1.76 & 2.74 & 2.67 \\
Ours & \underline{1.56} & \textbf{0.33} & \underline{1.34} & \textbf{1.08} \\ \bottomrule[1pt]
    \end{tabular}}
    \caption{Tracking Performance on TUM-RGBD~\cite{sturm12iros} (ATE RMSE [cm] $\downarrow$).} 
\label{tab:tum-t}
\end{table}

\subsection{Quality Of Reconstructed Map}

Table~\ref{tab:rep&sca} shows that PointSLAM++ significantly outperforms other methods in novel view rendering on RGB-D scenes across TUM-RGBD~\cite{sturm12iros} and Replica~\cite{replica19arxiv} datasets.

On Replica~\cite{replica19arxiv}, with its high-quality depth data and simpler scenes, PointSLAM++ achieves optimal results in all sequences.Among the comparative approaches, the state-of-the-art GS-ICP SLAM ranks second in rendering precision. PointSLAM++ successfully surpasses GS-ICP SLAM by leveraging neural Gaussian techniques. The implementation of a unique anchor point optimization technique skillfully addresses appearance variations, enabling PointSLAM++ to excel even under ideal conditions and outperform GS-ICP SLAM in rendering quality.

The TUM-RGBD dataset~\cite{sturm12iros} presents greater challenges with small objects, motion blur, and incomplete depth maps. Despite these difficulties, PointSLAM++ maintains the highest rendering accuracy across all sequences. GS-ICP SLAM performs poorly here due to its over-reliance on depth maps. PointSLAM++ uses depth primarily to enhance localization and geometry rather than letting depth noise directly affect rendering, significantly improving performance.

On ScanNet++~\cite{yeshwanth2023scannet++}, PointSLAM++ achieves a decisive victory where methods like GS-ICP SLAM fail to complete camera pose localization. PointSLAM++ enables high-quality map reconstruction and novel view rendering while maintaining stable localization, demonstrating its exceptional capability with challenging real-world scan data.

\subsection{Camera Tracking Accuracy}
Table~\ref{tab:rep&sca} shows our method's superior tracking accuracy on the Replica dataset, reducing trajectory errors by over 50\% compared to alternatives. By leveraging high-precision depth images and 3D information from G-ICP, we outperform methods relying on 2D spatial errors. Our performance is comparable to GS-ICP SLAM, as high-precision depth enables accurate camera pose predictions with GICP.

Table~\ref{tab:tum-t} shows our results on the TUM-RGBD dataset, our method achieves exceptional tracking accuracy through progressive pose optimization. By incorporating ORB features, we mitigate depth noise interference, enhancing localization stability. In contrast, GS-ICP SLAM performs poorly with noisy depth data, as its GICP algorithm struggles to deliver reliable results.

The evaluation results for the ScanNet++ dataset are listed in Table~\ref{tab:rep&sca}, our algorithm demonstrates tracking stability in scenarios where other methods fail. It addresses the dataset's challenges: large camera pose intervals, sparse textures, and depth interference. While GICP methods fail under these conditions and approaches relying on 2D spatial errors falter with substantial pose intervals, our method remains robust, showing its adaptability to complex environments. Additional results are in supply materials.

\begin{table}[t]
    \centering
    % \small
    \resizebox{\linewidth}{!}{
    \begin{tabular}{lcccc}
    \toprule[0.6pt]
       Method & Mapping Time$\downarrow$ & Tracking Time$\downarrow$ & FPS$\uparrow$ & GPU Memory(GB)$\downarrow$ \\
\midrule[0.5pt]
SplaTAM & 41m54s & 117m49s & 0.053 & 20.32 \\
MonoGS & 29m3s & 17m29s & 0.191 & 23.76 \\
GS-ICP SLAM & \textbf{1m10s} & 1m10s & 2.84 & \textbf{9.62} \\
Ours & 8m13s & \textbf{1m00s} & \textbf{3.33}  & 10.11 \\ \bottomrule[1pt]
    \end{tabular}}
    \caption{Comparison of Time Efficiency and Memory Consumption on the ScanNet++~\cite{yeshwanth2023scannet++}.} 
\label{tab:speed}
\end{table}

\subsection{Efficiency And Memory Comparison} 
Table~\ref{tab:speed} compares PointSLAM++'s efficiency and memory with existing methods on the ScanNet++~\cite{yeshwanth2023scannet++}. Compared to neural methods like SplaTAM and MonoGS, PointSLAM++ achieves high-quality reconstruction with significantly reduced processing time, due to our non-neural tracking optimization.

\begin{table}[t]
    \centering
    \small
    \resizebox{\linewidth}{!}{
    \begin{tabular}{lcccc}
    \toprule[0.6pt]
       Method & PSNR$\uparrow$ & SSIM$\uparrow$ & LPIPS$\downarrow$ & ATE RMSE$\downarrow$ \\
\midrule
w/o $PPO$ & 21.47 & 0.784 & 0.201 & 17.33 \\
w/o $NeuGS \& VDC$ & 24.06 & 0.812 & 0.168 & 2.32 \\
w/o $VDC$ & 25.36 & 0.856 & 0.128 & 1.10 \\
Ours & \textbf{26.16} & \textbf{0.885} & \textbf{0.106} & \textbf{1.08} \\ \bottomrule[1pt]
    \end{tabular}}
    \caption{Progressive Pose Optimization and Neural Gaussian Representation Ablation on TUM-RGBD~\cite{sturm12iros}.} 
\label{tab:ablation}
\end{table}

\subsection{Ablation Study}
We isolated two key modules from our progressive pose optimization framework and view-dependent environment compensation—and evaluated their impact through targeted experiments. Quantitative results are presented in Table~\ref{tab:ablation}.

\subsubsection{Progressive Pose Optimization Framework.} We compared the performance of PointSLAM++ with and without its progressive pose optimization (PPO). The results show a significant deterioration in tracking performance when relying solely on the GICP algorithm. This issue is pronounced under conditions with depth noise, where GICP alone leads to cumulative errors in geometrically ambiguous scenes. In contrast, the progressive approach effectively suppresses depth noise through its multi-stage adjustments. This method enhances the system's overall robustness and accuracy, proving beneficial in challenging scenarios involving rapid camera movement or dramatic changes in lighting.

\subsubsection{Neural Gaussian and View-Dependent Environment Compensation.} We established two ablation variants: one without neural Gaussians (NeuGS) and view-dependent compensation (VDC), and a second without only the VDC. The results show that incorporating neural Gaussians improves reconstruction quality over the baseline, particularly enhancing detail representation and geometric consistency. While this intermediate version captures richer geometric and appearance information, the full PointSLAM++ model demonstrates superior rendering consistency across different viewpoints and greater adaptability to lighting conditions. This confirms that the view-dependent compensation mechanism is crucial for addressing rendering artifacts caused by significant viewpoint changes, with improvements being evident in scenes with dramatic lighting variations. Furthermore, we also observed that although the tracking components are identical in the variants, the unique optimization process of the neural Gaussians leads to a more accurate map. This enhanced map fidelity, in turn, contributes to more precise pose estimation.
\section{Conclusion}
\label{sec:conclusion}
We propose PointSLAM++, a real-time RGB-D reconstruction system based on neural Gaussian representation. By integrating progressive pose optimization and view-dependent compensation, it mitigates depth noise, captures fine geometry and appearance, and ensures cross-view rendering consistency. Experiments demonstrate superior reconstruction accuracy and robustness for robotics, mixed reality, and intelligent interaction. Despite its effectiveness, the method increases computational cost, motivating future work on improving efficiency for broader device deployment.
% In this paper, we present PointSLAM++, a real-time RGBD reconstruction system leveraging neural Gaussian representation. Combining progressive pose optimization, neural Gaussian representation, and view-dependent environment compensation, our approach mitigates depth noise, captures detailed geometry and appearance, and resolves rendering inconsistencies across viewpoints. Extensive experiments show superior reconstruction quality, accurate camera poses, and robust performance, enabling reliable 3D reconstruction for robotics, mixed reality, and intelligent interaction. However, the enhanced representation raises computational complexity, challenging deployment on resource-limited devices. Future work will optimize efficiency to broaden device compatibility while preserving reconstruction quality.

\section{Acknowledgements}
This work was supported by the National Natural Science Foundation of China (No. U25A20421, No. 62202151, No. 62202152) and the National Key Research and Development Program of China (No. 2025YFB3003601).

\bibliography{aaai2026}

\end{document}